# Drawing on Memory: Dual-Trace Encoding Improves Cross-Session Recall in LLM Agents

Benjamin Stern
Tufts University
benjamin.stern@tufts.edu

Peter Nadel
Tufts University
peter.nadel@tufts.university

## Abstract

LLM agents with persistent memory store information as flat factual records, providing little context for temporal reasoning, change tracking, or cross-session aggregation. Inspired by the drawing effect [3], we introduce *dual-trace memory encoding*. In this method, each stored fact is paired with a concrete scene trace, a narrative reconstruction of the moment and context in which the information was learned. The agent is forced to commit to specific contextual details during encoding, creating richer, more distinctive memory traces. Using the LongMemEval-S benchmark (4,575 sessions, 100 recall questions), we compare dual-trace encoding against a fact-only control with matched coverage and format over 99 shared questions. Dual-trace achieves 73.7% overall accuracy versus 53.5%, a +20.2 percentage point (pp) gain (95% CI: [+12.1, +29.3], bootstrap $p < 0.0001$). Gains concentrate in temporal reasoning (+40pp), knowledge-update tracking (+25pp), and multi-session aggregation (+30pp), with no benefit for single-session retrieval, consistent with encoding specificity theory [8]. Token analysis shows dual-trace encoding achieves this gain at no additional cost. We additionally sketch an architectural design for adapting dual-trace encoding to coding agents, with preliminary pilot validation.

## 1 Introduction

Large language model agents with persistent memory can now maintain information across thousands of conversations, yet the way they store that information remains surprisingly flat. Whether using vector-indexed archival memory [5], retrieval-augmented generation over stored documents, or simple conversation-history concatenation, current systems record what was learned but discard when, where, and why it was learned. The result is a memory that functions like a box full of index cards: adequate for retrieving isolated facts, but unable to reconstruct temporal sequences, track how information changed over time, or synthesize details scattered across many encoding episodes.

This limitation matters most for the question types that distinguish a truly useful long-term assistant from a stateless lookup tool. When a user asks "what was the first problem with my car after its March service?" or "which social media platform grew fastest last month?", the agent must do more than locate a single relevant passage. It must sequence events, compare quantities across sessions, and distinguish updated information from stale records. These are precisely the capabilities where current flat-memory architectures fail.

We draw inspiration from cognitive research exploring the effect of drawing images on memory. Fernandes et al. [3] demonstrated that drawing a concept produces substantially stronger recall than writing it down, reading it, or even visualizing it. The critical mechanism is not motor activity or visual imagery, but *elaborative generation*, the act of producing a concrete, specific depiction, which forces the encoder to commit to particular spatial, temporal, and contextual details and creating a richer and more distinctive memory trace. Wammes et al. [9] showed that this benefit persists across retention intervals and encoding conditions, and is not attributable to longer study time or deeper semantic processing alone. The trace is stronger because it is more elaborated and thus connected to more contextual details that can later serve as additional retrieval cues.

We adapt this principle for LLM agents through **dual-trace memory encoding**. When our agent stores a piece of personal information, it creates two paired entries: a fact trace: a structured factual record, similar to what existing systems store, and a scene trace: a concrete, imageable narrative that embeds the same facts in a specific moment, place, and context. The scene trace forces the agent to perform elaborative generation at encoding time, committing to specific contextual details, for example, the visual setting, the temporal markers, the spatial relationships, rather than storing an abstract list of facts. At retrieval time, reconstructing the scene re-instantiates the encoding context, consistent with the encoding specificity principle [8].

We evaluate dual-trace encoding on LongMemEval-S [10], a benchmark of 4,575 real user conversation sessions with 100 structured recall questions spanning five memory capabilities. In a paired comparison over 99 shared questions between dual-trace encoding (C6-draw) and a fact-only control (C7) with comparable coverage and format, dual-trace achieves 73.7% overall accuracy versus 53.5%, a +20.2 percentage point gain (95% CI: [+12.1, +29.3], bootstrap $p < 0.0001$). The gain concentrates in temporal reasoning (+40pp), knowledge-update tracking (+25pp), and multi-session aggregation (+30pp), with zero benefit for single-session fact retrieval. This category-specific pattern is consistent with encoding specificity theory: scene traces help where episodic context creates additional retrieval pathways, and provide no benefit where a single fact suffices.

Our contributions are threefold:

1. We introduce a dual-trace memory encoding protocol for LLM agents, grounded in the drawing effect literature, with an evidence scoring gate that routes sessions to appropriate encoding categories and a three-state retrieval protocol.

2. We provide experimental evidence on LongMemEval-S showing that scene traces produce a +20.2pp accuracy gain over a fact-only baseline with comparable coverage and format across 99 shared questions ($p < 0.0001$). Category-level analysis reveals that gains are specific to temporal, aggregation, and update tasks, consistent with encoding specificity theory [8], with a clean null result on single-session retrieval.

3. We sketch an architectural design for coding adaptation, illustrating how dual-trace encoding could generalize beyond personal memory to domain-specific knowledge (debugging incidents, design decisions, learning progressions), with preliminary pilot validation on a Letta Code agent.

The remainder of this paper is organized as follows. Section 2 reviews related work on agent memory architectures and elaborative encoding in human cognition. Section 3 describes the dual-trace encoding protocol, evidence scoring, retrieval mechanism, experimental conditions, benchmark, and statistical methodology. Section 4 presents the experimental results. Section 5 sketches an adaptation to coding agents. Section 6 discusses the theoretical implications, iterative design lessons, and limitations. Section 7 concludes.

## 2 Related Work

### 2.1 Memory in LLM Agents

The problem of equipping LLM agents with persistent, cross-session memory has attracted substantial attention. We organize prior work by the memory architecture employed.

**Virtual context management.** MemGPT [5], later developed as the Letta framework, introduced a tiered memory architecture for LLM agents consisting of a core memory (editable persona and human blocks in the context window), a recall memory (searchable conversation history), and an archival memory (vector-indexed long-term storage). The agent manages its own memory through explicit tool calls, deciding what to store, retrieve, and update. This architecture provides the foundation for our work: our dual-trace protocol extends Letta's archival memory by pairing each stored fact with a scene trace.

**Observation-based memory.** Park et al. [7] introduced generative agents that maintain a stream of observations, periodically synthesizing higher-level reflections. Their retrieval mechanism scores memories by recency, importance, and relevance. While this approach captures temporal structure through the observation stream, it does not generate elaborative traces at encoding time; reflections summarize patterns rather than anchoring facts in concrete episodic contexts.

**Retrieval-augmented agent frameworks.** Retrieval-augmented generation (RAG) frameworks augment agent reasoning with context retrieved from external stores. These systems focus on retrieval architecture rather than encoding strategy: they optimize how stored information is found and injected into the context, but the stored representations themselves remain flat factual records.

**Memory benchmarks.** Two benchmarks have emerged for evaluating cross-session agent memory. LoCoMo [4] tests memory within a single long conversation, measuring an agent's ability to recall details from earlier in an extended dialogue. LongMemEval [10] tests memory across thousands of independent sessions, with structured questions spanning five capability types (single-session retrieval, multi-session aggregation, knowledge-update tracking, temporal reasoning, and abstention). We use LongMemEval-S, the single-user variant, because it exercises the cross-session encoding and retrieval that dual-trace encoding is designed to improve.

**Commercial memory layers.** Several commercial systems now provide memory persistence as a service, typically storing extracted facts or conversation summaries in vector databases. Mem0 [1], for instance, extracts factual records from conversations and stores them with vector embeddings for similarity-based retrieval. While effective for flat fact lookup, these systems focus on infrastructure, reliable storage, fast retrieval, API simplicity, rather than on the encoding representations themselves. No commercial memory layer generates elaborative scene traces at encoding time.

Across all of these approaches, the stored representation is a factual record: a summary, an extracted triple, or a raw observation. None generate elaborative scene traces at encoding time, and none leverage encoding-side generation as a mechanism for improving retrieval accuracy. Our work addresses this gap by showing that *how* information is encoded, not just *what* is stored or *how* it is retrieved, meaningfully affects downstream recall.

## 2.2 Elaborative Generation

Our approach draws on a line of research in human memory demonstrating that elaborative generation during encoding produces stronger, more retrievable memory traces.

Fernandes et al. [3] showed that drawing a word produces substantially better recall than writing it, reading it, visualizing it, or viewing a picture of it. The effect is robust across age groups, retention intervals, and experimental paradigms. Crucially, the benefit is not attributable to motor activity, longer study time, or deeper semantic processing alone. Wammes et al. [9] identified elaborative generation as the key mechanism: drawing forces the encoder to commit to specific spatial, structural, and contextual details, producing a trace that is both richer (more features encoded) and more distinctive (less confusable with other traces).

The drawing effect is consistent with the levels-of-processing framework [2], which predicts that deeper, more elaborative processing during encoding produces more durable memory traces. Scene generation in our protocol constitutes a deep processing operation: the agent must transform abstract facts into a concrete spatial-temporal narrative, engaging multiple representational dimensions.

Tulving and Thomson [8] established that retrieval is most effective when the cues available at retrieval match the context present at encoding. Our three-state retrieval protocol directly implements this principle: in State A, the agent reconstructs the scene before answering, re-instantiating the encoding context to maximize cue overlap.

Paivio [6] proposed that information encoded in both verbal and image formats is more retrievable than information encoded in one format alone, because it can be accessed through either representational system. Our fact + scene pairs create an analogous dual representation: the fact trace provides a structured verbal record, while the scene trace provides a concrete imagistic narrative. Although LLM agents do not have a separate visual processing system, the scene trace forces generation of spatial, temporal, and sensory details that create a functionally distinct representation from the fact trace.

The critical insight from this literature is that the benefit of drawing is not about drawing per se, it is about elaborative generation. Any encoding process that forces commitment to concrete, specific details should produce a similar benefit. LLM agents cannot draw, but they can generate concrete scene descriptions, and this generation process forces the same kind of elaborative commitment that makes drawing effective for humans.

### 2.3 Agent Memory for Software Engineering

Coding agents represent a natural extension domain for dual-trace encoding, since software engineering involves information types with strong temporal, rationale, and progression dimensions.

Letta Code provides a git-backed memory filesystem for coding agents, with structured initialization, explicit memory updates, and periodic memory check reminders. Facts are stored in category-specific blocks within the context window or as files in the memory filesystem. This architecture supports fact storage but does not include scene traces or elaborative encoding. Current coding assistants generally maintain project context within sessions but lack persistent cross-session memory architectures, relying on re-reading codebases and user-provided context rather than encoding and retrieving learned information across sessions. No current coding agent uses dual-trace or scene-based encoding for design decisions, debugging incidents, or developer learning progressions. In Section 5, we sketch an adaptation of dual-trace encoding for this domain.

## 3 Method

### 3.1 Dual-Trace Memory Encoding

We propose dual-trace encoding, a memory protocol in which every piece of personal information stored by an LLM agent consists of two paired entries: a fact trace and a scene trace (see Figure 1 for an overview).[1]

---

[1]The complete encoding protocol, including the evidence scoring rubric, scene quality rules, and retrieval protocol, is available as a deployable Letta agent skill at https://github.com/sternb12/agent_draw_skills.

The **fact trace** is a structured record tagged `[FACT:anchor]`, where `anchor` is a descriptive slug (e.g., `background_identity`, `car_maintenance_march`). It contains YAML frontmatter specifying the information type, category, confidence level, evidence score, and timestamp, followed by a `Components` list enumerating the specific details learned from the conversation. This is the format used by most existing agent memory systems and serves as our control representation.

The **scene trace** is tagged `[SCENE:anchor]` with the same anchor slug, linking it to its paired fact. It begins with the word "Picture:" followed by a concrete, imageable narrative that embeds the same factual details into a specific moment and spatial context. For example, rather than storing the fact that a user raised $200 at a church bake sale and completed a 5K charity run in 35 minutes, the scene trace describes a corkboard with a race bib reading "Finished: 35:00" pinned next to a church bulletin with "$200 raised" circled. Each scene trace concludes with the disclaimer "(Mnemonic depiction only. Not evidence.)" to prevent downstream confusion between reconstructed scenes and verified facts.

The theoretical motivation comes from the drawing effect [3]. In human memory experiments, drawing a concept produces substantially stronger recall than writing, reading, or even mental visualization. The mechanism is elaborative generation: the act of producing a concrete, specific depiction forces the encoder to commit to particular details, the spatial layout, the objects present, the temporal markers, creating a richer, more distinctive trace with multiple potential retrieval cues. Our scene traces perform the same elaborative function: the agent must generate specific visual-spatial details during encoding, not merely copy a factual list. This is consistent with deeper levels of processing [2] and with dual coding theory [6], which predicts that information encoded in both verbal and image formats is more retrievable than information encoded in only one format.

The fact and scene traces are cross-linked via `linked_scene` and `linked_fact` fields in their respective frontmatter, ensuring that retrieval of either trace surfaces the other.

## 3.2 Evidence Scoring

Not every conversation warrants dual-trace encoding. Most interactions with an LLM assistant are task-oriented requests (e.g., "summarize this article," "write a regex for email validation") that contain no personal information worth storing. To avoid noise accumulation, we introduce an evidence scoring gate that evaluates each session before encoding.

The scoring system uses three dimensions, each rated 0–2. First, relevance measures whether the user explicitly states personal facts (0 = task context only, 1 = incidental personal context, 2 = explicit personal disclosure). Next, specificity captures whether the information is concrete enough to be recalled later (0 = vague, 1 = general, 2 = specific with names, numbers, dates, or events). Last, explicitness reflects how directly the information was stated (0 = implied, 1 = casual mention, 2 = direct statement). The three scores sum to a total between 0 and 6.

The scoring system evolved across experimental conditions. In the initial evidence-scored condition (C4), three routing tiers determined encoding depth: sessions scoring 0–2 were classified as DROP (no entries created), sessions scoring 3–4 as STREAMLINED (fact trace only), and sessions scoring 5–6 as FULL (both fact and scene traces), with a stakes override forcing FULL encoding at score $\geq 3$ for discrete items with real-world retrieval consequences. In the controlled comparison conditions (C6-draw and C7-control), routing was simplified to two tiers: DROP (0–2) and FULL (3–6), eliminating the STREAMLINED tier. This change increased session coverage from approximately 22% (C4) to approximately 55–57% (C6/C7). On the LME-S benchmark, approximately 78% of sessions are correctly classified as DROP across all conditions, containing no personal information worth storing.

## 3.3 Three-State Retrieval Protocol

At retrieval time, the agent follows a three-state protocol that calibrates its confidence based on what memory traces it finds.

In **State A**, where fact and scene both is found, the agent first reconstructs the scene internally, then answers the question with high confidence. The scene reconstruction re-instantiates the encoding context, consistent with the encoding specificity principle [8]. For aggregation queries that span multiple topics, the agent searches across entries, reads all matching fact and scene pairs, uses temporal anchors from the scenes to sequence events, and synthesizes a composite answer.

In **State B**, where a fact, but no scene is found, the agent answers from the fact trace alone with medium confidence. It does not fabricate or hallucinate a scene. This state arises when a session was encoded with a fact trace but no accompanying scene, either because the condition does not generate scenes (as in C7-control) or because the encoding did not produce a scene for that particular session.

In **State C**, where nothing is found, the agent explicitly abstains: "I don't have that information stored." This calibrated abstention is critical for trustworthy memory systems: the agent should not confabulate answers from general knowledge when the question asks about the user's specific experiences.

## 3.4 Experimental Conditions

We evaluate dual-trace encoding through a controlled comparison between two conditions that isolate the contribution of scene traces.

**C6-draw** (the dual-trace condition) uses the simplified two-tier gate (DROP/FULL at the 0–2/3–6 threshold), producing paired `[FACT:anchor]` and `[SCENE:anchor]` entries for all sessions that pass the evidence gate. Both C6 and C7 store memory as structured archival entries that the agent searches via its retrieval tools. C6 processed 4,375 of the 4,575 LME-S sessions, achieving 54.8% session coverage (2,399 sessions where at least one entry was created), with 2,726 total memory inserts. Of the stored sessions, 92.0% (2,208) received dual-trace encoding and 8.0% (191) received fact-only encoding.

**C7-control** (the fact-only condition) uses the same evidence scoring gate, the same `[FACT:anchor]` naming format, and the same routing logic, but generates no scene traces. C7 processed all 4,575 sessions, achieving 57.4% session coverage (2,627 sessions stored). All stored entries used the `[FACT:anchor]` format without scene traces.

The two conditions achieve comparable coverage (54.8% vs. 57.4%), ensuring that differences in recall accuracy reflect the scene trace contribution rather than differences in how much information was stored. What C7 controls for is critical to the experimental design: both conditions share the same format (structured `[FACT:anchor]` traces with YAML frontmatter), the same naming convention (descriptive anchor slugs), and the same evidence gate threshold. What C7 isolates is the scene trace.

Both conditions were run on the same infrastructure (Letta API, Claude Sonnet as the agent model, 300-second timeout per session). The C6 teach phase completed in 4.1 hours at an average rate of 17.6 sessions per minute with no restarts. The C7 teach phase ran at a comparable rate but required multiple restarts due to intermittent API stalls in the final 200 sessions; an automated monitor script detected rate stalls and resumed from the last completed session, ensuring no data loss. Recall evaluation for each condition processed 100 questions in approximately 20 minutes.

For context, we also report results from six additional conditions representing the developmental trajectory of the system: vanilla Letta (no memory protocol), C1 (an early dual-trace attempt with a broken persona-based trigger), C2 (fact-only with low coverage), C3 (improved baseline), C4 (evidence-scored encoding with lower coverage), and C5-FS (C4 plus a file-system index for retrieval). These conditions are not the focus of the controlled comparison but illustrate the iterative refinement path.

### 3.5 Benchmark: LongMemEval-S

We evaluate on LongMemEval-S (LME-S), the single-user variant of the LongMemEval benchmark [10]. LME-S consists of 4,575 real user conversation sessions drawn from the ShareGPT corpus, paired with 100 structured recall questions and ground-truth oracle answers.

The 100 questions span four question types, each testing a different memory capability: single-session fact retrieval (20 questions), multi-session aggregation (20 questions), knowledge-update tracking (20 questions), and temporal reasoning (20 questions). An additional 20 abstention questions are distributed across these four types, testing whether the agent correctly declines to answer when the relevant information was never mentioned. Following the benchmark's evaluation protocol, we report abstention accuracy as a separate category. All per-type accuracy figures use the 20 non-abstention questions per type as the denominator, with the 20 abstention questions pooled into a fifth "abstention" category.

All conditions use the same LME-S sessions and the same 100 recall questions. The evaluation pipeline uses GPT-4o as a judge to compare each agent response against the ground-truth oracle answer, following LongMemEval's `evaluate_qa.py` script. C6-draw was graded on 99 of 100 questions (one abstention question was excluded from the C6 evaluation due to a processing artifact); all paired comparisons between C6 and C7 use the 99 questions common to both.

### 3.6 Statistical Methodology

We use two complementary statistical approaches. For the primary hypothesis test (C6 accuracy > C7 accuracy), we compute a one-sided bootstrap $p$-value using 10,000 resamples of the 99 paired question outcomes, with a fixed random seed of 42 for reproducibility. The null hypothesis is that the accuracy difference between C6 and C7 is zero or negative.

For all accuracy estimates and differences, we report 95% bootstrap confidence intervals computed from the same 10,000 resamples. Per-category intervals are computed by resampling within each category ($n = 20$ questions per type, except $n = 19$ for abstention in C6).

For the per-question agreement analysis, we use McNemar's test with continuity correction, which is the appropriate test for paired binary outcomes. We report the chi-squared statistic and its associated $p$-value.

## 4 Results

### 4.1 Overall Accuracy

Table 1 presents the accuracy of all eight experimental conditions on LME-S, broken down by question type.

The vanilla Letta agent, which has no memory encoding protocol, achieves 20.0% overall, correctly abstaining on all 20 abstention questions but answering none of the 80 non-abstention questions correctly. This baseline confirms that without an explicit encoding protocol, Letta's default memory architecture retains no retrievable personal information across 4,575 sessions. The developmental conditions (C1 through C5-FS) show a gradual improvement from 38.4% to 48.5%, reflecting iterative refinements in encoding coverage, format, and retrieval architecture. C7-control, the fact-only condition with high coverage and clean formatting, reaches 53.5%. C6-draw, the dual-trace condition, achieves 73.7%, the highest accuracy across all conditions by a substantial margin.

Table 1: Accuracy (%) on LME-S by condition and question type. Per-type columns report accuracy on the 20 non-abstention questions per type; Abst reports accuracy on the 20 pooled abstention questions. Overall is computed across all 100 questions (or 99 for C6).

| Condition | Overall | Single | Multi | Know-upd | Temporal | Abst |
|---|---|---|---|---|---|---|
| Vanilla Letta | 20.0 | 0 | 0 | 0 | 0 | 100 |
| C1 (broken dual-trace) | 40.4 | 50 | 0 | 35 | 20 | 100 |
| C2 (fact-only) | 38.4 | 40 | 15 | 30 | 20 | 85 |
| C3 (baseline) | 47.5 | 55 | 5 | 55 | 30 | 95 |
| C4 (evidence-scored) | 47.5 | 60 | 10 | 45 | 30 | 95 |
| C5-FS (+index) | 48.5 | 60 | 20 | 45 | 25 | 95 |
| C7-control (fact-only) | 53.5 | 75 | 20 | 55 | 25 | 95 |
| **C6-draw (dual-trace)** | **73.7** | **75** | **50** | **80** | **65** | **100** |

Table 2: C6 vs. C7 accuracy by question type with 95% bootstrap CIs (10,000 resamples). Delta is the paired difference in percentage points. $p$-values are one-sided bootstrap tests of C6 > C7. The paired comparison uses the 99 questions common to both conditions; abstention $n = 19$ for C6.

| Category | C6-draw | C7-control | $\Delta$ | 95% CI | $p$ |
|---|---|---|---|---|---|
| Single-session | 75 [55, 90] | 75 [55, 90] | 0 | $[-15, +15]$ | 0.657 |
| Multi-session | 50 [30, 70] | 20 [5, 40] | +30 | $[+10, +50]$ | 0.001 |
| Knowledge-update | 80 [60, 95] | 55 [35, 75] | +25 | $[+10, +45]$ | 0.003 |
| Temporal reasoning | 65 [45, 85] | 25 [10, 45] | +40 | $[+15, +65]$ | 0.002 |
| Abstention | 100 [100, 100] | 95 [84, 100] | +5 | $[0, +16]$ | 0.351 |
| **Overall** | **73.7** [65, 82] | **53.5** [43, 64] | **+20.2** | $[+12, +29]$ | $< 0.0001$ |

### 4.2 The Scene Trace Signal: C6 vs. C7

The controlled comparison between C6-draw and C7-control isolates the contribution of scene traces, since the two conditions differ only in whether facts are paired with scenes. Figure 2 and Table 2 present the category-level differences with 95% bootstrap confidence intervals.

Overall, C6-draw outperforms C7-control by 20.2 percentage points (73.7% vs. 53.5%), with a 95% confidence interval of $[+12.1, +29.3]$ that excludes zero. The bootstrap $p$-value is less than 0.0001: none of 10,000 resamples produced a difference of zero or less, indicating that the scene trace contribution is robust to sampling variability.

The category-level pattern is theoretically informative. Scene traces produce no gain on single-session fact retrieval (75% vs. 75%, $\Delta = 0$pp, $p = 0.657$). Both agents retrieve single-session facts with equal precision when only one passage needs to be found. The gain is concentrated in the three temporally complex categories: multi-session aggregation (+30pp, $p = 0.001$), knowledge-update tracking (+25pp, $p = 0.003$), and temporal reasoning (+40pp, $p = 0.002$). Abstention is stable across both conditions (100% vs. 95%, $p = 0.351$), confirming that scene traces do not cause the agent to over-answer questions about information it never stored.

## 4.3 Per-Question Agreement Analysis

To examine the scene trace effect at the individual question level, we construct the $2 \times 2$ agreement table between C6 and C7 on the 99 questions common to both conditions.

Of the 99 questions, 51 were answered correctly by both agents, 24 were missed by both, 22 were answered correctly by C6 but missed by C7, and only 2 were answered correctly by C7 but missed by C6. McNemar's test with continuity correction yields $\chi^2 = 15.04$, $p < 0.001$, confirming that the asymmetry between the 22 C6-only wins and the 2 C7-only wins is highly significant.

The 22 questions where scene traces made the difference break down by category: 9 temporal-reasoning, 6 multi-session, 5 knowledge-update, 1 single-session, and 1 abstention. This concentration in temporally complex question types is consistent with the theoretical prediction that scene traces provide retrieval value specifically for tasks requiring temporal sequencing, change detection, and cross-session synthesis.

## 4.4 Interpretation

Five findings emerge from these results.

**Coverage and format produce modest gains.** The progression from C4 (evidence-scored, 8.7% coverage) to C7 (fact-only, 57.4% coverage) represents a +6pp improvement. Encoding more sessions with cleaner labels and higher coverage helps, but the effect is small relative to the scene trace contribution. This result suggests that encoding depth, what gets stored alongside each fact, matters more than encoding breadth.

**Scene traces add a further +20.2pp.** This is the isolated scene contribution, with coverage and format held constant between C6 and C7. We see that the elaborative generation mechanism from the drawing effect literature [3] transfers to LLM agent memory at scale.

**The single-session null result is theoretically predicted.** When a single passage suffices to answer a question, both agents retrieve it with equal success (75% vs. 75%). Scene traces provide no additional retrieval pathway in this regime. The gain is specific to tasks that require distinguishing, sequencing, or synthesizing memories across multiple encoding episodes, exactly the regime where encoding specificity theory [8] predicts that richer encoding context should improve retrieval.

**Statistical significance is strong across temporally complex categories.** Multi-session aggregation ($p = 0.001$), knowledge-update tracking ($p = 0.003$), and temporal reasoning ($p = 0.002$) all show individually significant gains. The overall McNemar's test ($p < 0.001$) and bootstrap test ($p < 0.0001$) confirm that the pattern is not attributable to chance.

**Abstention stability is preserved.** Both C6 and C7 show high abstention accuracy (100% vs. 95%, $p = 0.351$). The stable abstention rate confirms that scene traces improve recall without degrading the agent's ability to recognize when it lacks relevant information.

## 4.5 Token Efficiency

A practical concern with dual-trace encoding is cost. We analyze token usage from the experimental runs.

In the teach phase, C6-draw consumed an average of 156,430 tokens per session (155,800 prompt + 631 completion). C7-control consumed 159,211 tokens per session (158,935 prompt + 275 completion). Despite generating 2.3× more completion tokens per session, C6 was 1.7% cheaper overall. The scene-writing overhead of approximately 356 additional completion tokens per session is negligible relative to the prompt tokens that dominate session cost (∼156,000 tokens per session). Then in the recall phase, C6-draw consumed 244,385 tokens per query (474 completion), while C7-control consumed 252,750 tokens per query (284 completion). C6 was 3.3% cheaper per query despite generating 67% more completion tokens. Dual-trace encoding does not increase cost. The practical implication of this token-level analysis is that dual-trace encoding delivers a 20-percentage-point accuracy gain at no additional token cost.

Importatnly, all conditions used Claude Sonnet 4.6 (`anthropic/claude-sonnet-4-6`) as the agent model and `text-embedding-3-small` (OpenAI) as the embedding model. C6 teach phase: 4,375 sessions in 4.1 hours (17.6 sessions/min, 0 restarts). C7 teach phase: 4,575 sessions in ∼4.2 hours (multiple restarts due to rate stalls on the final 200 sessions).

# 5 Extension: Coding Agents

Software engineering presents a natural extension domain: design decisions carry rationale that may be forgotten, debugging incidents unfold as temporal narratives, and developer skills evolve over time. These are information types where our results predict scene traces should add retrieval value. We sketch an architectural adaptation for Letta Code agents; controlled experimental evaluation is left for future work.[2]

The adaptation targets Letta Code, a coding agent framework built on Letta/MemGPT with a git-backed memory filesystem. The design principle is to extend, not replace: dual-trace adds a scene layer on top of existing fact storage, preserving compatibility with existing workflows.

Two key changes distinguish the coding variant. First, evidence scoring expands from three conversational dimensions (each 0–2, total 0–6) to four code-specific dimensions (each 0–3, total 0–12): durability, scope, rationale-richness, and retrieval likelihood. Routing uses three tiers: SKIP (0–4, native memory only), RECORD (5–7, fact entry), and FULL (8–12, fact+scene). Second, the scene vocabulary shifts from "Picture:" (visual-spatial metaphors for personal experiences) to "Moment:" (narrative reconstructions anchored to concrete code artifacts such as file paths, function names, and error messages). Each coding scene includes Timeline, Prior, and After fields for temporal sequencing. The adaptation supports six information types, decisions, incidents, conventions, patterns, learning progressions, and preferences, each with domain-appropriate fact and scene representations.

We conducted four manual pilot tests on a Letta Code agent (March 2026). The evidence gate correctly discriminated low-score preferences (SKIP) from high-score architectural decisions (FULL). State A retrieval and State C abstention functioned as designed, and an update to a previously stored incident correctly modified the existing entries rather than creating duplicates. One notable finding: a system directive instructing the agent to load the dual-trace skill was reliably overridden by the agent's trained default behavior of writing to its native memory block. The encoding protocol had to be specified inline in the system prompt to take effect, suggesting that agent memory protocols competing with trained defaults require sufficient prompt-level priority.

---

[2]The full design, including the scoring rubric, scene vocabulary, information-type specifications, and pilot test transcripts, is published as a Letta Code agent skill at https://github.com/sternb12/letta-code-draw-skill.

# 6 Discussion

## 6.1 Why Scenes Help: Elaborative Encoding, Not Visual Imagery

The +20.2pp accuracy gain from dual-trace encoding raises an obvious question: why do scene traces improve retrieval? Our results are most consistent with an elaborative encoding explanation rather than a visual imagery explanation.

The scene generation process forces the agent to commit to specific contextual details during encoding, specifying the spatial layout of objects, the temporal markers linking events, and the sensory anchors distinguishing one memory from another. This is the same mechanism that makes drawing effective in human memory experiments [3]. The benefit comes from the forced commitment to concrete detail, not from the modality of the output. An LLM agent has no visual processing system, yet the act of generating a concrete scene description produces a functionally richer trace than a factual list.

Scene traces also create multiple retrieval pathways. A fact trace can be retrieved via semantic similarity to the query. A scene trace adds episodic retrieval cues, including temporal markers, spatial anchors, and relational details that match different aspects of a query. This is consistent with dual coding theory [6]. Information encoded in two functionally distinct representations is more retrievable than information encoded in one.

The single-session null result (0pp difference, zero discordant questions) provides important evidence. When only one passage needs to be found and a single fact suffices to answer the question, scene traces provide no additional retrieval pathway. The gain appears only when the task requires distinguishing, sequencing, or synthesizing information across multiple encoding episodes. This pattern indicates that scene traces improve memory by creating distinctive contextual anchors that help differentiate and organize memories, not by making individual facts easier to find.

The temporal reasoning results suggest that scene traces enable active self-correction during retrieval. In one case, the agent was asked which of two hobby projects was started first. The agent initially retrieved the wrong project and stated it was first. However, the scene trace for that project contained a concrete date anchor and a detail that the other project was "already on a shelf nearby, partially assembled from a few weeks before." Cross-referencing these temporal anchors, the agent caught its own error and self-corrected in real time. In the fact-only condition, the agent lacks these concrete temporal anchors and has no pathway for detecting its own reasoning errors. We suggest that scene traces may actively support error detection and recovery during multi-step retrieval reasoning. We note this as a qualitative observation from a single case, not as systematic evidence for a self-correction mechanism. Whether this pattern holds across the full set of discordant questions remains to be established.

## 6.2 Encoding Specificity in Agent Memory

Tulving and Thomson's (1973) encoding specificity principle states that retrieval is most effective when the cues present at retrieval match the context present at encoding. Our three-state retrieval protocol implements this principle. In State A, the agent reconstructs the scene before answering, re-instantiating the encoding context.

The question of where the scene trace contributes, at encoding, at retrieval, or at both, remains open. The current design couples scene generation (encoding) with scene reconstruction (retrieval). An ablation experiment that generates scenes during encoding but withholds them during retrieval would isolate the encoding-side contribution. Conversely, an ablation that provides scene traces at retrieval for entries that were encoded without them would test whether the retrieval-side reconstruction alone accounts for the benefit. Although we did not run these ablations, we note this as a future research opportunity.

### 6.3 The Iterative Path: Lessons from C1 through C5

The development path from C1 to C7 reveals which design decisions matter for agent memory and which do not. We highlight three lessons.

First, C1's failure demonstrates that encoding triggers must match evaluation framing. C1 used a persona-based trigger that did not align with the evaluation protocol's question framing. The agent stored information but could not retrieve it under the evaluation conditions. We concluded that memory encoding and retrieval should be designed as a coupled system, not as independent components.

Second, the modest progression from C1 through C5 (40.4% to 48.5%) shows that incremental improvements to encoding coverage, naming conventions, and retrieval indexing produce limited gains when the fundamental encoding representation remains flat. The jump from C5 (48.5%) to C7 (53.5%) via higher coverage and cleaner formatting adds only 5pp. The jump from C7 (53.5%) to C6 (73.7%) via scene traces adds 20.2pp. To us, this suggests that encoding depth overshadows encoding breadth.

Third, the shift in retrieval strategy was the critical architectural change that enabled the C6/C7 gains. Earlier conditions (C1–C5) relied primarily on embedding-similarity search over archival memory, which works well for semantic matching but poorly for temporal reasoning and aggregation because it returns isolated passages without structural context. C6 and C7 use structured entries that the agent can search, read in full, and reason over explicitly, cross-referencing temporal anchors across entries rather than depending on a single best-match.

### 6.4 Limitations

Several limitations qualify the interpretation of our results.

We evaluate on a single benchmark (LongMemEval-S). While LME-S provides a rigorous, standardized evaluation with realistic session data and diverse question types, additional benchmarks and real-world deployments are needed to establish generalizability.

A GPT-4o evaluator grades all answers, introducing evaluator variance. We follow the LongMemEval evaluation protocol without modification, but automated evaluators can disagree with human judgments on borderline cases.

Per-category sample sizes are small ($n = 20$ questions per type), resulting in wide confidence intervals at the category level. While all three temporally complex categories show confidence intervals excluding zero, a larger question set would provide more precise estimates of category-specific effect sizes.

The encoding-retrieval confound discussed in Section 6.2 means we cannot attribute the gain exclusively to encoding-side or retrieval-side mechanisms. Both may contribute, and their relative contributions remain unmeasured.

The coding adaptation presented in Section 5 is an architectural design with preliminary pilot validation, not a controlled experiment.

Finally, our bootstrap confidence intervals assume question-level independence. In practice, some LME-S questions share topical overlap, which may introduce mild dependence that our bootstrap procedure does not account for.

## 6.5   Generalizability

The dual-trace principle, pairing a factual record with a concrete contextual scene, is domain-agnostic. The coding adaptation in Section 5 demonstrates one domain transfer, adapting the scene vocabulary and evidence scoring to software engineering information types. The same approach could be applied to medical agents (pairing clinical facts with patient encounter narratives), legal agents (pairing case details with contextual decision scenes), or educational agents (pairing learned concepts with specific teaching moments).

The key requirement is that the domain includes information types with temporal, rationale, or progression dimensions, the cases where scene traces add retrieval value. Domains consisting primarily of static, context-free facts would not benefit from dual-trace encoding, consistent with our single-session null result.

# 7   Conclusion

This paper introduced dual-trace memory encoding for LLM agents, a protocol inspired by the drawing effect in human cognition that pairs each stored fact with a concrete scene trace. On LongMemEval-S, dual-trace encoding achieves 73.7% overall accuracy versus 53.5% for a fact-only control with matched coverage and format, a +20.2 percentage point gain (95% CI: [+12.1, +29.3], bootstrap $p < 0.0001$). The gain concentrates in temporal reasoning (+40pp), knowledge-update tracking (+25pp), and multi-session aggregation (+30pp), with a clean null result on single-session fact retrieval. This category-specific pattern is consistent with the hypothesis that scene traces improve cross-session recall through elaborative encoding and contextual anchoring, not simply via storage expansion.

Per-question analysis confirms the robustness of the effect: 22 questions were answered correctly by the dual-trace agent but missed by the fact-only control, versus only 2 in the reverse direction (McNemar's $\chi^2 = 15.04$, $p < 0.001$). Token analysis shows that dual-trace encoding achieves this gain at no additional cost: encoding is 1.7% cheaper per session and retrieval is 3.3% cheaper per query than the fact-only control.

We have also sketched an adaptation of dual-trace encoding for coding agents, with preliminary pilot validation showing successful routing, retrieval, abstention, and update behaviors.

Future work should pursue three directions: first, an encoding-retrieval ablation to isolate whether the benefit comes from scene generation at encoding time, scene reconstruction at retrieval time, or both; second, real-world deployment studies to evaluate dual-trace encoding outside of benchmark settings; and third, controlled experimental validation of the coding adaptation with a domain-specific benchmark.

## Code Availability

The complete dual-trace encoding protocol for personal memory, including the evidence scoring rubric, three-state retrieval protocol, worked examples from the LME-S evaluation, and token analysis, is publicly available as a deployable Letta agent skill at: https://github.com/sternb12/agent_draw_skills (MIT License). The coding adaptation described in Section 5, including the four-dimension scoring rubric, six information types, scene format, and the four pilot test transcripts referenced in that section, is published separately as a companion Letta Code skill at: https://github.com/sternb12/letta-code-draw-skill (MIT License).

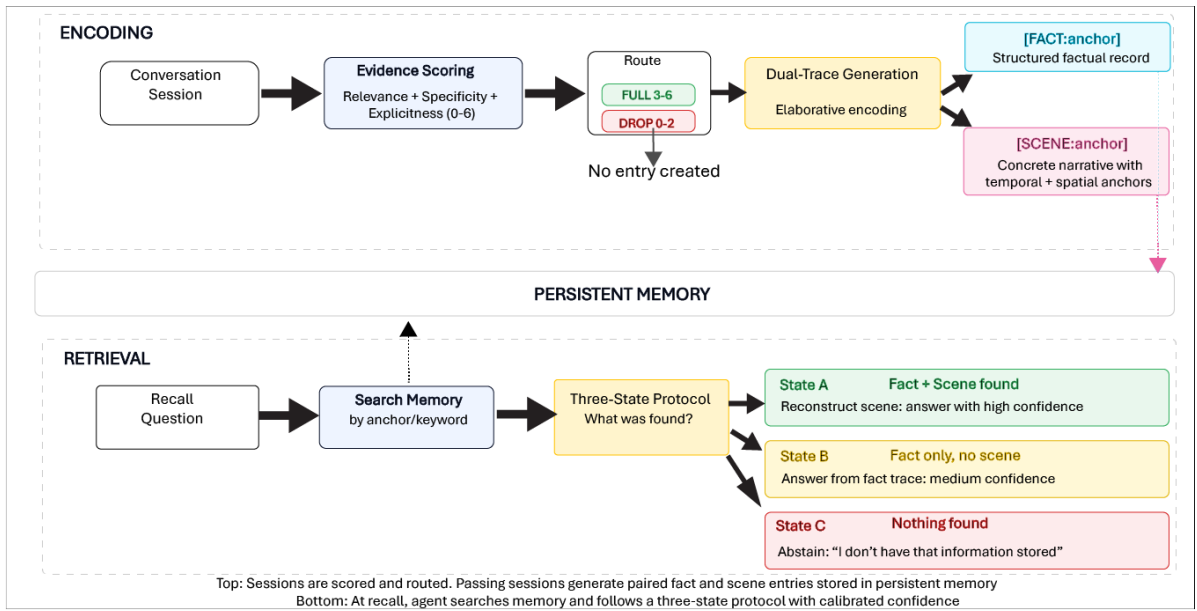

Figure 1: Overview of the dual-trace encoding and retrieval protocol. **Encoding** (top): each session is scored on three evidence dimensions (Relevance, Specificity, Explicitness; each 0–2, total 0–6). Sessions scoring 0–2 are dropped; sessions scoring 3–6 receive full dual-trace encoding, producing a linked [FACT:anchor] and [SCENE:anchor] pair stored in persistent memory. **Retrieval** (bottom): the agent searches memory and follows a three-state protocol: State A (fact + scene found, high confidence), State B (fact only, medium confidence), or State C (nothing found, explicit abstention).

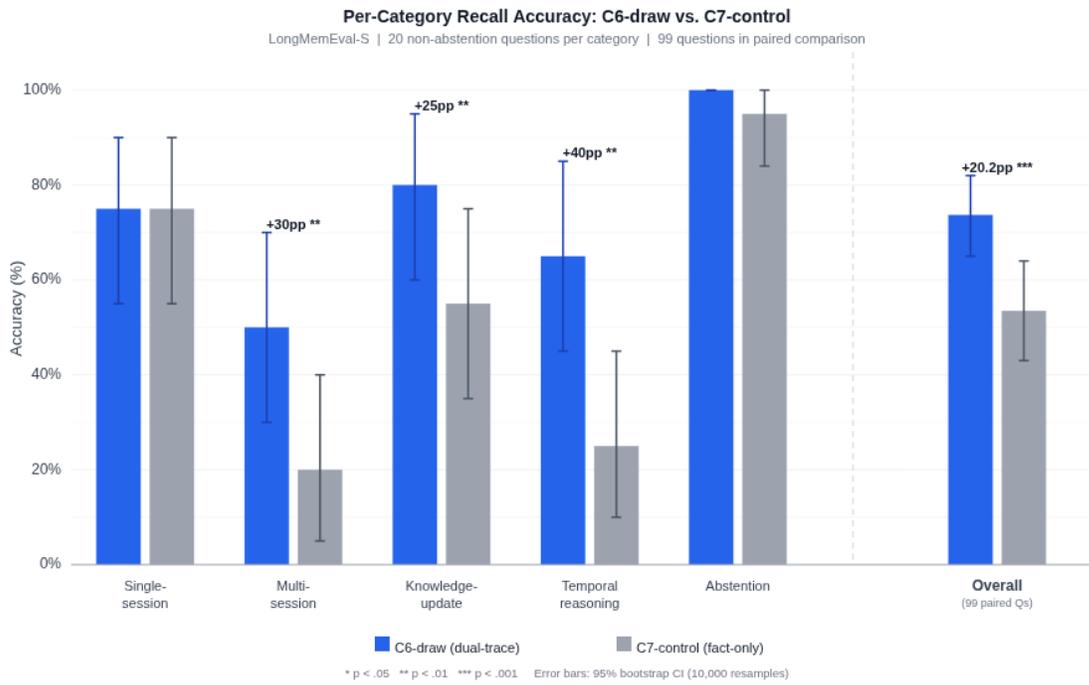

Figure 2: C6-draw (dual-trace) vs. C7-control (fact-only) accuracy on LongMemEval-S by question category. Delta labels show the percentage point gain for categories with statistically significant differences. The overall comparison uses the 99 questions common to both conditions.